\documentclass[utf8]{frontiersSCNS} 

\usepackage{url,hyperref,lineno,microtype,subcaption}
\usepackage[onehalfspacing]{setspace}
\usepackage[ruled,vlined]{algorithm2e}
\usepackage{amsmath}

\def\keyFont{\fontsize{8}{11}\helveticabold }
\def\firstAuthorLast{I. Magrans, R. Kanai}
\def\Authors{Ildefons Magrans de Abril\,$^{1,*}$ and Ryota Kanai\,$^{1}$}
\def\Address{ARAYA, Inc., Tokyo, Japan}
\def\corrAuthor{Corresponding Author}

\def\corrEmail{ildefons.magrans@gmail.com}

\begin{document}
\onecolumn
\firstpage{1}

\title[A unified strategy for implementing curiosity and empowerment driven RL]{A unified strategy for implementing curiosity and empowerment driven reinforcement learning} 

\author[\firstAuthorLast ]{\Authors} 
\address{} 
\correspondance{} 

\extraAuth{}

\maketitle

\begin{abstract}

\section{}

Although there are many approaches to implement intrinsically motivated artificial agents, the combined usage of multiple intrinsic drives remains still a relatively unexplored research area. Specifically, we hypothesize that a mechanism capable of quantifying and controlling the evolution of the information flow between the agent and the environment could be the fundamental component for implementing a higher degree of autonomy into artificial intelligent agents. This paper propose a unified strategy for implementing two semantically orthogonal intrinsic motivations: curiosity and empowerment. Curiosity reward informs the agent about the relevance of a recent agent action, whereas empowerment is implemented as the opposite information flow from the agent to the environment that quantifies the agent's potential of controlling its own future. We show that an additional homeostatic drive is derived from the curiosity reward, which generalizes and enhances the information gain of a classical curious/heterostatic reinforcement learning agent. We show how a shared internal model by curiosity and empowerment facilitates a more efficient training of the empowerment function. Finally, we discuss future directions for further leveraging the interplay between these two intrinsic rewards.

\tiny
 \keyFont{ \section{Keywords:} intrinsic motivation, reinforcement learning, curiosity, empowerment, homeostasis} 
\end{abstract}

\section{Introduction}



Within a reinforcement learning setting \citep{sutton1998reinforcement}, a reward signal indicates a particular momentary positive (or negative) event and it serves to constrain the long-term agent behavior. Extrinsic rewards are generated by an external oracle and they indicate how well the agent is interacting with the environment (e.g. videogame score, portfolio return). On the other hand, intrinsic rewards are generated by the agent itself and they indicate a particular internal event sometimes implemented as a metaphor of an animal internal drive \citep{chentanez2005intrinsically,barto2004intrinsically,sequeira2011emotion,song2006stimulating}.


There are many intrinsic rewards and most of them can be characterized by how they affect the information flow between the environment and the agent. In one side of the spectrum, information is pushed from the agent to the environment, for instance, by rewarding actions that lead to predictable consecutive sensor readings \citep{montufar2016information} or by rewarding reaching states from where the agent actions have a large influence in determining the future state (i.e. empowerment \citep{jung2011empowerment,mohamed2015variational,karl2017unsupervised, gregor2016variational}).


On the other side, information is encouraged to efficiently move from the environment to the agent. These rewards motivate the agent to explore its environment by taking actions leading to an improvement of its internal models. \cite{jurgen1991possibility} proposed an online learning agent equipped with a curiosity unit measuring the Euclidean distance between the observed state and the model prediction. Recently, \cite{pathak2017curiosity} extended the curiosity functionality to accommodate agents with high dimensional sensory inputs by adding a representation network to filter out information from the observed state not relevant for predicting how the agents actions affect the future state. \citep{houthooft2016vime} presented an exploration reward bonus based on information gain maximization computed using a variational approximation of a Bayesian neural network. \cite{lopes2012exploration} discussed an exploration reward bonus that encourages the learning progress over the last few experiences instead of the immediate agent surprise. \cite{bellemare2016unifying} differ in the sense that the agent is not learning a forward model but a probability density function about the states visited by the agent together with a lower bound on  the information gain associated with the agent exploratory behavior.

Although there are many approaches to implement intrinsically motivated artificial agents, the combined usage of multiple intrinsic drives is still a relatively unexplored research field. To establish a principled approach to combine multiple types of intrinsic motivations, we propose an approach in which an agent is designed to optimize the information flow between the agent itself and the environment. By learning to sense and act  via the internal representations of the information flow, the agent would be able to behave as if it were rewarded by a particular intrinsic reward function. As we discuss below, this general formulation of intrinsic motivation can capture a large spectrum of emergent autonomous behaviors, from a curious agent aiming to acquire as much information as possible to an agent aiming to reach highly empowering states. With this architecture, an agent could discover new curiosity-driven behaviors by simply vising new internal states or sequences of internal states. We believe that the generality of this internal representation, independent of a particular task and/or agent sensing/acting capabilities, has the potential to foster multi-agent and multi-task architectures with new transfer learning capabilities.

This paper is our first step towards developing an intelligent agent capable of sensing and acting according to the information flow between the environment and itself. Our contribution in this paper is our proposal for an implementation method to compute the state of the information flow. It quantifies the information gain and empowerment obtained by an agent interacting with the environment at every step. In the following section we discuss our design requirements and present our approach. Section 3 presents the experimental results. Finally, the discussion section summarizes our main results and limitations along with possible future directions.

\section{Background}

\begin{figure}[h!]
  \centering
  \includegraphics[width=0.80\textwidth]{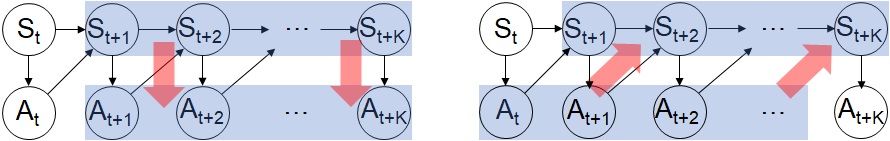}
  {\caption{Conceptual diagram of the information gain process (left) and empowerment (right): dark thin arrows are causal dependencies and large arrows show the direction of the information flow. }
  \label{fig:diagram}}
\end{figure}

This paper assumes a typical reinforcement learning setup where an agent interacts with the environment at discrete time steps, it observes a state $s_t$ $\in$ $S$ and it acts on the environment with action $a_t$ $\in$ $A$ according to a control policy $a_t \sim \pi(A_t|s_t)$. Within this setting, \cite{tiomkin2017unified} presented recursive expressions to describe the information transferred from a sequence of environment states to the sequence of agent actions as well as to describe the information transferred from the agent actions to the environment states. In both cases, it is assumed that the agent interacts open-endedly with a Markovian environment (i.e.  transition probability function $\sim$ $P(S_{t+1}|S_t,A_t)$). Figure \ref{fig:diagram} shows two different points of views of the information flow for the same process of an agent interacting with a Markovian environment. Equations \ref{eqn:infoenvagent} and \ref{eqn:infoagentenv} present the recursive expressions of the information transferred from environment to agent and from agent to environment respectively:

\begin{eqnarray}
  \label{eqn:infoenvagent}
   && InfoEnvToAgent(s_t)=I(S_{t+1:t+K} \rightarrow A_{t+1:t+K} || S_{t:t+K-1},A_t) \nonumber \\
  &&=I(S_{t+1};A_{t+1}|S_{t},A{t}) \nonumber \\
  &&+<I(S_{t+2:t+K} \rightarrow A_{t+2:t+K} || S_{t+1:t+K-1},A_{t+1})>_{P(S_{t+1},A_{t+1}|S_t,A_t)} ,
\end{eqnarray}

\begin{eqnarray}
  \label{eqn:infoagentenv}
   && InfoAgentToEnv(s_t)=I(A_{t:t+K-1} \rightarrow S_{t+1:t+K}|s_t) \nonumber \\
  &&=I(A_t;S_{t+1}|s_t)+<I(A_{t+1:t+K} \rightarrow S_{t+2:t+K}|S_{t+1})>_{P(S_{t+1}|s_t)} ,
\end{eqnarray} 
where lower case is used for concrete states and actions, uppercase is used to denote random variables, $A$/$S_{t+1:t+K}$ is the sequence of actions/states of length $K$ starting at time $t+1$ and $I(X_{1:N} \rightarrow Y_{1:N} || C_{1:N})$ is the causally conditioned directed mutual information \citep{kramer1998directed}:

\begin{eqnarray}
  \label{eqn:causallyconditioneddefinition2}
   && I(X_{1:N} \rightarrow Y_{1:N} \lVert Z_{1:N})=\sum_{i=1}^{N} I(X_{1:i};Y_i|Y_{1:i-1}Z_{1:i}) 
\end{eqnarray}  
where this definition differs from that of the conditional mutual information only on that $X_{1:i}$ and $Z_{1:i}$ substitutes $X_{1:N}$ and $Z_{1:N}$. The causal conditioning $||$ reflects a causal relationship on past and present only.

It is important for our approach that both equations have a recursive structure decomposition similar to the Bellman equation. From this point of view, $I(S_{t+1};A_{t+1}|S_{t},A{t})$ and $I(S_{t+1};A_{t}|s_{t})$ would act as agent reward when we try to encourage our agent to take actions that maximize the information flows from the environment to the agent and from the agent to the environment respectively. 

As we discussed in the introduction, intrinsic rewards can be characterized by how they affect the information flow between the environment and the agent. Then, we should be able to encourage a variety of behaviors, similar to those encouraged by particular intrinsic rewards, by properly balancing the two types of rewards derived from equations (\ref{eqn:infoenvagent}) and (\ref{eqn:infoagentenv}). Therefore, if we could create an agent that can sense and act in a space that quantifies the strength of the information flow between itself and the environment, then we should be able to enhance the agent capacity to thrive on different, previously unknown environments by controlling its movement in this internal space. 


These rewards define our internal space. Computing these rewards requires the computations of the corresponding conditional mutual information which requires the approximation of the corresponding probability distributions \citep{mohamed2015variational,tiomkin2017unified}. When actions and/or states are discrete, we can approximate them for instance using a neural network with a softmax output layer. However it’s much harder when states and actions are continuous, especially when the state space is very high dimensional (e.g. video stream). In the following sections, we propose a more practical method to implement both rewards.


\section{Curiosity with homeostatic regulation}
\label{sec:curiosity}

This section discusses a practical method to compute the reward coming from equation (\ref{eqn:infoenvagent}) defined as $I(S_{t+1};A_{t+1}|S_{t},A{t})$. Our method avoids the approximation of complex distributions over continuous states and actions. We validate this first reward function using a state of the art RL algorithm that works well with continuous actions. We chose the Deep Deterministic Policy Gradient algorithm \citep{lillicrap2015continuous} but other options are also feasible. This algorithm finds a deterministic control policy that maximize the expected sum of discounted rewards. When $\gamma = 1$, episode length is $K$ and reward function is $I(S_{t+1};A_{t+1}|S_{t},A{t})$, then our agent explores the environment by maximizing the information gain as expressed in equation (\ref{eqn:infoenvagent}).

\begin{figure}[h!]
  \centering
  \includegraphics[width=0.90\textwidth]{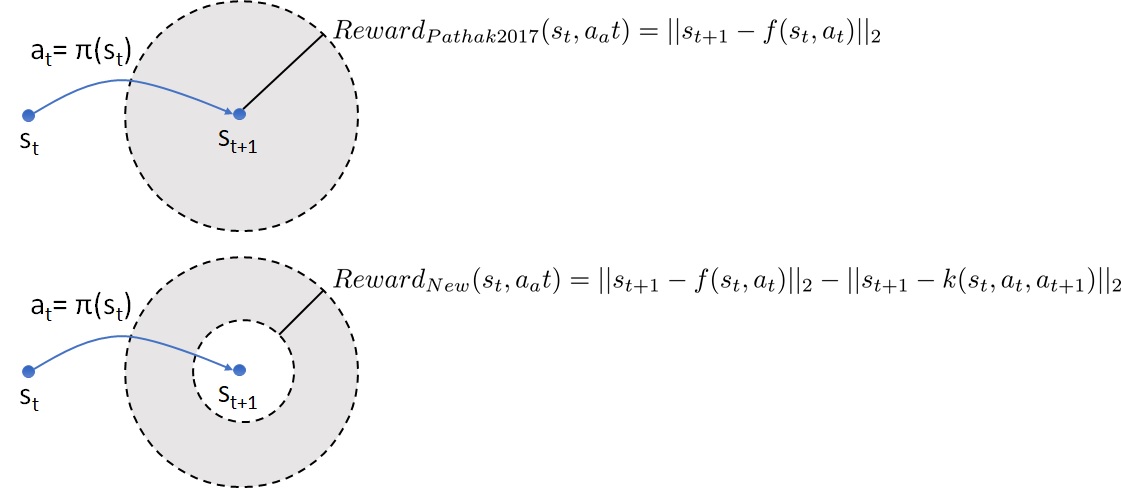}
  {\caption{Semantic of the curiosity reward with homeostatic regulation and comparisson with respect to a state of the art curiosity reward based on the Euclidean distance between the observed state and the model prediction (e.g. \citep{jurgen1991possibility,pathak2017curiosity}).}
  \label{fig:curiositySemantic}}
\end{figure}

We can express this reward as the reduction of entropy in the future state $S_{t+1}$. Then, because we are able to know exactly the current state and due to the deterministic nature of the control policy inferred by the DDPG algorithm, we use the concrete state $s_t$ and actions $a_t$ and $a_{t+1}$ instead of the random variables $S_t$, $A_t$ and $A_{t+1}$ respectively to compute the reward. Finally, we approximate the reduction of entropy in the future state $S_{t+1}$ as the reduction of the prediction error in the future state. Equation \ref{eqn:simplification} formalizes this approximation:

\begin{eqnarray}
  \label{eqn:simplification}
   && I(S_{t+1};A_{t+1}|S_{t},A_{t})= H(S_{t+1}|S_{t},A_{t}) - H(S_{t+1}|S_{t},A_{t},A_{t+1}) \nonumber \\
   && \approx H(S_{t+1}|s_{t},a_{t}) - H(S_{t+1}|s_{t},a_{t},a_{t+1}) \nonumber \\
  && \approx ||s_{t+1} -\hat{s}_f||_2-||s_{t+1} -\hat{s}_k||_2
\end{eqnarray} 
where $\hat{s}_f = f(s_t,\pi(s_{t})=a_t)$ and $\hat{s}_k = k(s_t,\pi(s_{t})=a_t,\pi(s_{t+1})=a_{t+1})$ are the future state predictions by the forward and extended forward models respectively. The extended forward model takes advantage of the knowledge of the action that the agent will take in the future state to improve the prediction about this future state. This approximation captures the relevant semantic with much lower computational cost. Interestingly, the internal models $f(.)$ and $k(.)$ can be easily implemented with deep neural networks, which can accomodate an agent with high-dimensional input streams. Figure \ref{fig:curiositySemantic} is a graphical representation of the semantic of the new curiosity reward and how it compares with respect to a state of the art curiosity reward based on the Euclidean distance between the observed state and the model prediction (e.g. \citep{jurgen1991possibility,pathak2017curiosity}).

Our new curiosity reward has two components: 1) Heterostatic motivation: similarly to a state of the art work based on the Euclidean distance \citep{jurgen1991possibility,pathak2017curiosity}, the first component of our reward encourages taking actions that lead to large forward model errors. This first component implements the heterostatic drive. In other words, the tendency to push away our agent from a predictable behavior; 2) Homeostatic motivation: the second component is our novel contribution. It encourages taking actions $a_t$ that lead to future states $s_{t+1}$ where the corresponding future action $a_{t+1}$ gives us additional information about $s_{t+1}$. This situation happens when the agent is ``familiar'' with the state-action pair: $(s_{t+1},a_{t+1})$. Therefore, our new reward encourage the agent to move towards regions of the state-action space that simultaneously deliver large forward model errors and that are ``known/familiar'' to the agent. In other words it implies a priority sampling strategy towards ``hard-to-learn'' regions of the state-action space. 

We further generalize this reward by adding an hyper-parameter $\alpha>0$ that controls the importance of the of the homeostatic bonus. It is interesting to note that this reward is equal to the curiosity reward proposed by \citep{pathak2017curiosity} when $\alpha = 0$. Finally, we should note that the reward function is non-stationary due to the continuous learning of $f$ and $k$. For that reason we $z$-normalize the reward using a mean and standard deviation computed at the end of each of episode using all available samples:

\begin{eqnarray}
  \label{eqn:finalreward1}
   && R(s_t)=  \frac{IG_{\alpha}(s_t)-\mu_{ig}}{\sigma_{ig}}
\nonumber \\
   && IG_{\alpha}(s_t)=||s_{t+1} -\hat{s}_f||_2-\alpha||s_{t+1} -\hat{s}_k||_2
\end{eqnarray} 
where $\mu_{ig}$ and $\sigma_{ig}$ are the sample mean and sample standard deviation of the reward computed according to all samples collected so far.  Algorithm \ref{alg:main} summarizes the overall logic of our curiosity agent. It follows an architecture similar to \cite{pathak2017curiosity}:

\begin{algorithm}
\SetAlgoLined
\KwResult{Forward model: $f(s,a)$ }
 $N$: Total number of training episodes\; 
 $K$: Duration of each exploration episode\;
 Initialization of $f, k, DDPG$ parameters including $\pi$\;
 Initialization of random exploration probability $\epsilon$\;
 \For{episode i:1..N}{
 Initialize environment: initial state $s_0$ according to experiment strategy (see section \ref{results})\;
 \For{step t:1..K}{
 	Generate $a_t=\pi(s_t)$ (random according to $\epsilon$)\;
 	Sample $s_{t+1} \sim P(.|st, at)$\;
 	Get reward $r_t$ according to equation (\ref{eqn:finalreward1})\;
 	Add $\{s_t,a_t,s_{t+1},r_t\} $ to Replay Buffer (RB)\;
 	Sample Mini-Batch $MB \sim RB$\;
 	Train internal models $f, k$ and DDPG networks (e.g. $\pi$) using $MB$;\
 	
  }
 }
 
\caption{Curiosity-driven reinforcement learning with homeostatic regulation}
\label{alg:main}
\end{algorithm}

\section{Approximated empowerment}

This section discusses the implementation of the reward coming from equation (\ref{eqn:infoagentenv}) defined as $I(A_t;S_{t+1}|s_t)$. We validate the implementation of this second reward by fitting a deterministic control policy with the DDPG algorithm that is able to guide an agent following the path of the maximum empowerment. We implement the definition of empowerment proposed by \citep{tiomkin2017unified}:

\begin{eqnarray}
\label{eqn:empowermenttiomkin}
   && Empowerment(s_t)=max_{\mathbf{\omega}}\;I(A_{t:t+K-1} \rightarrow S_{t+1:t+K}|s_t) \nonumber \\
  &&=max_{\mathbf{\omega}}\;I(A_t;S_{t+1}|s_t)+<I(A_{t+1:t+K} \rightarrow S_{t+2:t+K}|S_{t+1})>_{P(S_{t+1}|s_t)} 
\end{eqnarray}
where $\omega(A_t|s_t)$ is the source distribution. This definition of empowerment is based on the mutual information betweeen a sequence of actions and the corresponding sequence of future states which is slightly different than the original empowerment definition by \cite{klyubin2005empowerment} which is based on the mutual information between a sequence of actions and the final state after executing all actions:

\begin{eqnarray}
\label{eqn:empowermentoriginal}
   && Empowerment(s_t)=max_{\mathbf{\omega}}\;I(A_{t:t+K-1};S_{t+K}|s_t) 
\end{eqnarray}


Similarly to previous section, we take advantage of the Bellman like equation of the information transferred from the agent to environment (Eq. \ref{eqn:infoagentenv}) to justify the use of a reinforcement learning algorithm which finds a control policy that maximizes $Empowerment(s_t)$ over a sequence of $K$ steps. Crucially, we assume that the reward function is stationary and known before we start optimizing the control policy. Therefore, we could compute it using dynamic programming. However we will continue using DDPG algorithm to stress the similarities with the curiosity-driven agent, presented in previous section, and the potential interplay between the models required to compute both rewards.

In this case, the reward function at state $s_t$ is defined by $R_{emp}(s_t)=max_{\mathbf{\omega}}\;I(A_t;S_{t+1}|s_t)$. A key additional cost of computing this reward, compared with the reward discussed in previous section, is that we have to optimize the source distribution $\omega$ that delivers the maximum conditional mutual information. To address the high computational cost of this reward, we perform a number of approximations. We express the mutual information as the reduction of entropy in the future state: $I(A_{t:t+K-1};S_{t+K}|s_t)=H(S_{t+1} | s_t)-H(S_{t+1} | A_t , s_t)$. The first approximation step is to compute the first entropy term $H(S_{t+1} | s_t)$ assuming a fixed uniform source distribution $a_t\sim\omega(A_t|S_t)=U(min,max)$ instead of optimizing $\omega$ as in the original formulation. This reward component implements a measure of future possible states according to a fixed uniform source distribution. The second entropy term, defined by $H(S_{t+1} | A_t , s_t)$, is approximated using only the action provided by the deterministic control policy. In other words, we assume that the actions are distributed according to a Dirac delta distribution optimized by the DDPG algorithm: $a_t\sim\delta_{\pi(s_t)}$. This second entropy term would capture the agent potential to move from the current state to the future state in a controlled way. Finally, we approximate the first and second entropy terms respectively as follows:

\begin{eqnarray}
  \label{eqn:entroptyterm1}
   && H(S_{t+1}|s_{t}) \approx \frac{1}{N}\sum_{a_i\sim U(.)} ||\hat{s}_{t+1}-f(s_t,a_i) ||_2\nonumber \\
   && \hat{s}_{t+1} = \frac{1}{N}\sum_{a_i\sim U(.)} f(s_t,a_i) \nonumber \\
  && H(S_{t+1}|A_t,s_{t})\approx ||s_{t+1} - f(s_t, a_t=\pi(s_t))||_2
\end{eqnarray} 
where $f$ is the forward model, $U(.)$ is the uniform distribution in the action space and $\pi(s_t)$ is the deterministic control policy at time $t$. As it has been discussed in this section, we assume an stationary reward. Therefore, according to equation (\ref{eqn:entroptyterm1}), we are assuming that the forward model is known. 

\begin{figure}[h!]
  \centering
  \includegraphics[width=0.8\textwidth]{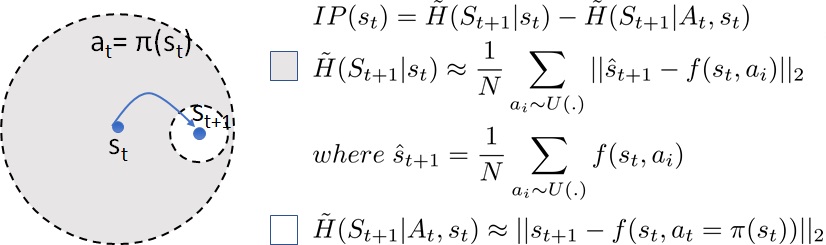}
  {\caption{semantic of the new approximated empowerment reward. Light grey area represents the area of possible future states $s_{t+1}$ assuming a uniform distribution of actions. White area is the possible deviation from the future state predicted by the forward model when the agent takes the action suggested the current control policy $\pi(s_t)$.Therefore, this reward encourages policies that lead the agent towards states with large number of future possibilities and 
states from where the future state is highly predictable given the action defined by the control policy.
}
  \label{fig:empowermentSemantic}}
\end{figure}

Figure \ref{fig:empowermentSemantic} is a graphical representation of the semantic of the new approximated empowerment reward discussed in this section. Light grey area represents the area of possible future states $s_{t+1}$ assuming a uniform distribution of actions. White area is the possible deviation from the future state predicted by the forward model when the agent takes the action suggested the current control policy $\pi(s_t)$. Therefore, this reward encourages policies that lead the agent towards states with large number of future possibilities and states from where the future state is highly predictable given the action defined by the control policy.

This reward captures the semantic of the original reward defined in the trivial term of equation (\ref{eqn:empowermenttiomkin}), it avoids the maximization over the source distribution and the approximation of complex distributions over states and actions. The assumption of having a forward model is actually a feature rather than a limitation because the forward model becomes the main instrument of interplay between the curiosity and the empowerment internal functions. A first example of this interplay is discussed in section \ref{results}, where we show that we can efficiently train our forward model using the curiosity agent and then use this same model to compute the reward of a second empowerment-driven agent.  More details of this interplay will be presented in a future paper. Not optimizing $\omega$, could indeed be a more important limitation specially when the environment is not deterministic and the probabilistic response $p(S_{t+1}|s_t,a_t)$ is not uniform across the state-action space. The final approximation step in both rewards presented in equations (\ref{eqn:simplification}) and (\ref{eqn:empowermenttiomkin}) are based on the $L_2$ norm which is not a valid distance metric when the state space is not Euclidean. \cite{pathak2017curiosity} showed that this limitation can be solved by fitting a representation network $\hat{s}=\phi(s)$ using the reconstruction performance of an agent action decoder as loss function.

\section{Results}
\label{results}

\subsection{Curiosity: experiment 1}
\label{curiosityExp1}
Our experimental validation presents two examples where both curiosity and homeostatic drives are superior to learn a forward model. Our validation hypothesis is that exploring an environment with several non-linearities could be optimized by regulating the agent curiosity with a homeostatic drive. More specifically, it prioritize the exploration of the state-action space according to how hard it is to learn.    

\begin{figure}[h!]
  \centering
  \includegraphics[width=0.35\textwidth]{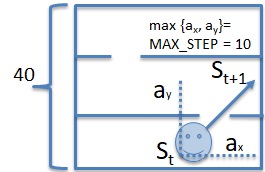}
  {\caption{Scheme of our 3 room environment.}
  \label{fig:diagram3}}
\end{figure}

To test our hypothesis, we use a 3 room continuous space environment of 40 by 40, where an agent, able to sense its exact position, learns a control policy according to the DDPG algorithm with the reward presented in equation (\ref{eqn:finalreward1}) and a probability of taking a random action equal to $0.5$. The available actions are bi-dimensional action vectors such that $max(a_x,a_y)=10$. The environment is deterministic and when an agent collides with a wall it returns to its previous state. The agent starts every episode in a random state and it runs for 10 steps (with max length step=10). We have implemented the forward model $f$ and the extended forward model $k$ as feed forward neural networks with 2 hidden layers with 64 hidden units each. We store the agent traces and we train the agent and the internal models at the same time following algorithm \ref{alg:main}.  Figure \ref{fig:diagram3} shows a scheme of our environment. 

\begin{figure*}
  \centering
  \includegraphics[width=0.80\textwidth]{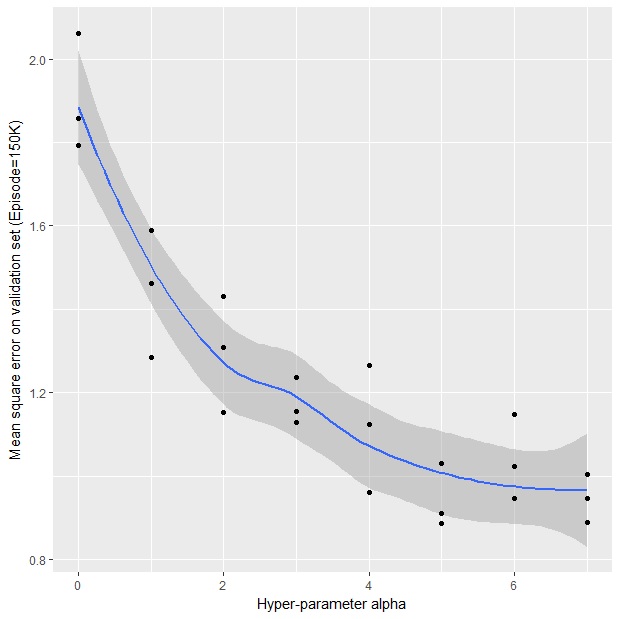}
  {\caption{Accuracy of the forward model learned by the agent as a function of $\alpha$ (measured according to the mean square error on the validation set).}
  \label{fig:resultsexperiment1}}
\end{figure*}

\begin{figure*}
  \centering
  \includegraphics[width=0.8\textwidth]{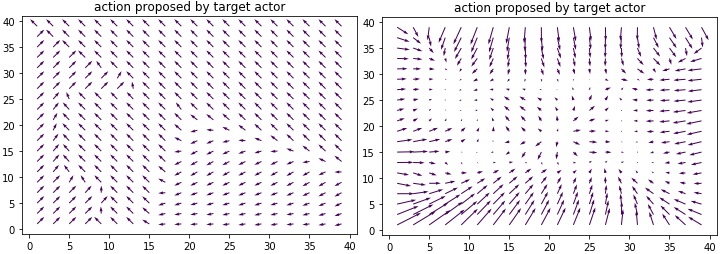}
  {\caption{Flow diagram of the control policy learned after 10K episodes with $\alpha=0$ (left) and $\alpha=7$ (right) respectively.}
  \label{fig:resultsexperiment1b}}
\end{figure*}

In our first experiment we study the accuracy of the final forward model as a function of $\alpha$. We check the prediction accuracy using a validation data set of $10^7$ randomly generated samples collected  independently of the training process and never used to train $f$ and/or $k$. We run our agent using different values of $\alpha \in \{0,1,2,3,4,5,6,7\}$  for $150000$ episodes and we do each experiment 3 times. Figure \ref{fig:resultsexperiment1} shows how we can improve the environment sampling efficiency by increasing the homeostatic component of the reward (i.e. $\alpha$). Figure \ref{fig:resultsexperiment1b} shows a diagram of the policy learned after $10000$ episodes with $\alpha=0$ and $\alpha=7$ respectively. We can clearly appreciate that, when $\alpha$ is large, the agent tends to position itself where there are a larger number of non-linearities (i.e. the ``doors''). This agent behavior enhances the learning of complex regions by leveraging a more intense random exploration where it is most required. 

We should also mention, that for this particular experiment, a pure random sampling strategy achieves a mean square error, on the validation set, of 0.67 which is better than the best result obtained with $\alpha=7$ (0.87). However this is not a fair comparison because every episode starts in a different position which enables a pure random agent to reach every spot of the environment by simply random walking its local surroundings while our curiosity agent is constrained by the a relatively low random exploration probability ($0.5$). For instance, we are able to beat the random sampling agent performance using our agent with a random exploration probability equal to $0.9$ and $\alpha=7$. In this case, we achieve an average mean square error over three runs of $0.55$ which is better than the $0.67$ achieved by the random sampling agent.

\subsection{Curiosity: experiment 2}
\label{curiosityExp2}

\begin{figure*}
  \centering
  \includegraphics[width=0.80\textwidth]{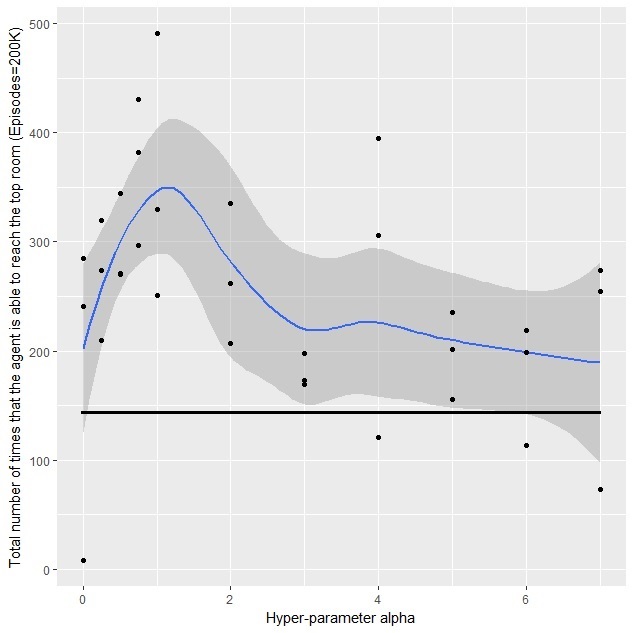}
  {\caption{Total number of times that the agent is able to reach the top room as a function of $\alpha$ when it starts every episode in a random position of the bottom room.}
  \label{fig:resultsexperiment2}}
\end{figure*}

We performed a second experiment using the same environment described in Figure \ref{fig:diagram3}, but in this case the agent starts every episode in a random state of the bottom room. We want to understand whether the homeostatic reward is able to enhance the acquisition of innovative environment samples by counting how many times the agent is able to traverse 2 doors and reach the top room. Figure \ref{fig:resultsexperiment2} shows how we can improve the acquisition of challenging environment states by optimizing the contribution of the homeostatic reward component. In this case, a pure random sampling strategy running for $150000$ episodes only reach the top room a total average of 145 times which is far below any other total average achieved with a non-random strategy with any $\alpha$.  

\subsection{Empowerment: experiment 1}

\begin{figure*}
  \centering
  \includegraphics[width=0.80\textwidth]{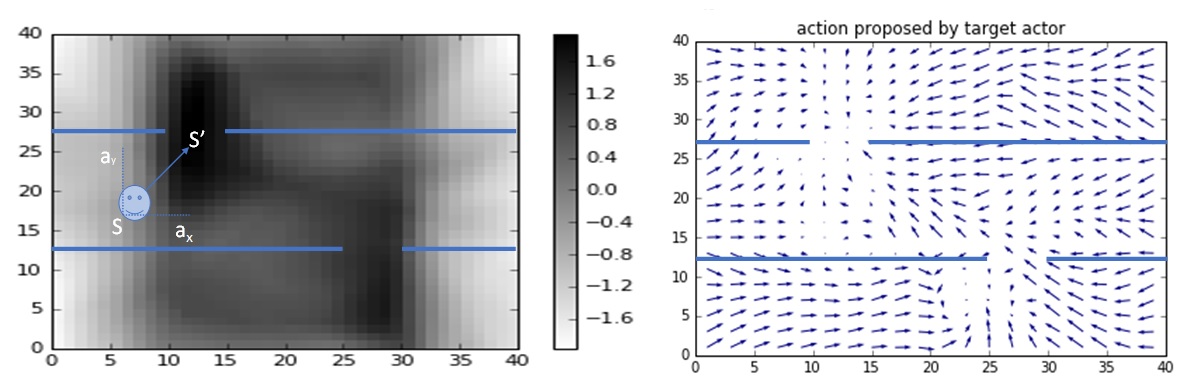}
  {\caption{Agent environment and approximated empowerment reward profile (left). Control policy to maximize the acquisition of reward (right).}
  \label{fig:empowerAgent}}
\end{figure*}

To test our approximation of empowerment, we use again the 3 room environment where an agent, able to sense its exact position, starts every episode in a random state and it runs for 10 steps (with max length step=10). We have implemented the reward function defined in equation (\ref{eqn:entroptyterm1}) with a forward model $f$ trained using the curiosity agent described in section \ref{curiosityExp1} with a random exploration probability equal to $0.9$ and $\alpha=7$. With the forward model completely trained, we optimize our control policy using DDPG algorithm. Figure \ref{fig:empowerAgent} shows a diagram of the reward function (left) and the final control policy (right). Our empowerment approximation rewards the agent to position itself close to the apartment doors because this position provides the larger amount of future options to the agent.  

\section{Discussion}

We presented a new approach to define the internal state space of a learning agent. Our strategy is to create a minimal set of internal functions that summarize the state of the information flow between the agent and the environment. In this paper, we proposed a unified framework for implementing two types of intrinsic motivations, namely, curiosity and empowerment from the perspective of information flow between the agent and the environment. Curiosity was implemented as the drive to increase information flow from the environment to the agent whereas empowerment was formulated as the information flow from the agent to the environment. With these unified intrinsic motivations, we hypothesized that an agent should be able to generate a broad spectrum of autonomous behavior

The curiosity function quantifies interestingness of a particular state-action pair, while the empowerment function measures the agent future options and control at the current state. These are computed at every discrete time step using two functions that depend of the actual observations, agent actions and internal forward model of the environment. We derive them from information theoretical considerations and proposed methods to minimize the computational cost and share internal models across the two types of intrinsic motivations. 
 
The curiosity function quantifies two opposing animal drives: 1) the innate drive to explore (heterostatic behavior) and 2) the desire to maintain certain critical parameters stable. We presented an exploration approach to demonstrate this first function. It generalizes a state of the art method \citep{pathak2017curiosity} and we present experimental results to demonstrate the superior exploration behavior of our joint homeostatic and heterostatic drive with respect to a pure curiosity/heterostatic approach. 

The second derived function (i.e. empowerment) quantifies at each state the trade-off between the amount of possible future states assuming a uniform distribution of one step actions, and the precision to move to the next state according to a deterministic control policy. This function is our proposal to quantify the information that an agent can transmit to the environment when following a deterministic control policy and it is based on similar information theory principles as the curiosity function. We evaluated this function by optimizing a control policy that follows a sequence of states that are the optimal trade-off between amount of possible future states reachable from each state and control accuracy. This example captures the semantic of an empowerment-driven agent at a much lower computational cost than the original formulation.

In the future work, we will explore meta-learning strategies to dynamically adjust the contribution of the two intrinsic reward functions as well as the homeostatic drive (i.e. $\alpha$), the random exploration probability and eventually also an external reward. Our meta-controller should be able to compute a probabilistic model over bi-dimensional functions on a space defined by the weights of the intrinsic rewards. Gaussian processes are good candidates as they offer a sample efficient way to approximate this distribution as well as principled approaches to implement the intrinsic weight sampling strategies \citep{snoek2012practical}. It is key to address the non-stationary behavior of both intrinsic motivation functions \citep{snoek2014input}.

\section*{Conflict of Interest Statement}

The authors were employed by company ARAYA, Inc.

\section*{Author Contributions}

IM and RK conceived the method. IM performed experiments and analyzed data. IM and RK wrote the paper.

\section*{Funding}
This work was supported by JST CREST Grant Number JPMJCR15E2, Japan.




\bibliographystyle{frontiersinSCNS_ENG_HUMS} 
\bibliography{frontiers}





\end{document}